\documentclass[letterpaper, 10 pt, conference]{ieeeconf}  

\usepackage{color,soul}
\usepackage{cite}
\usepackage{amsmath,amssymb,amsfonts}
\usepackage{algorithmic}
\usepackage{graphicx}
\usepackage{textcomp}
\usepackage{comment}
\usepackage{tabu}    
\usepackage{booktabs}
\usepackage{multirow}
\usepackage{multicol}
\usepackage{url}
\usepackage{fancyhdr}

\IEEEoverridecommandlockouts                              

\title{\LARGE \bf
Deep Predictive Learning with Proprioceptive and Visual Attention for Humanoid Robot Repositioning Assistance 
}

\author{Tamon Miyake$^{1}$, Namiko Saito $^{1, 2}$, Tetsuya Ogata $^{3, 4}$, Yushi Wang$^{1}$, and Shigeki Sugano $^{3}$
\thanks{*This work was supported by JST Moonshot R and D, Grant No. JPMJMS2031.}
\thanks{$^{1}$Tamon Miyake, Namiko Saito and Yushi Wang are with Future Robotics Organization, Waseda University, Tokyo, Japan. (Corresponding author: Tamon Miyake (e-mail: tamonmiyake$@$aoni.waseda.jp)) $^{2}$Namiko Saito is with Microsoft Research Asia, Tokyo, Japan.}
\thanks{$^{3}$Tetsuya Ogata and Shigeki Sugano are with Faculty of Science and Engineering, Waseda University, Tokyo, Japan. $^{4}$Tetsuya Ogata is with The National Institute of Advanced Science and Technology, Tokyo, Japan.}
}

\fancypagestyle{firstpage}{
  \fancyfoot[C]{\footnotesize The 2025 IEEE/RSJ International Conference on Intelligent Robots and Systems (IROS 2025), Hangzhou, China}
}

\begin{document}
\maketitle
\thispagestyle{empty}
\pagestyle{empty}

\begin{abstract}

Caregiving is a vital role for domestic robots, especially the repositioning care has immense societal value, critically improving the health and quality of life of individuals with limited mobility.
However, repositioning task is a challenging area of research, as it requires robots to adapt their motions while interacting flexibly with patients. 
The task involves several key challenges: (1) applying appropriate force to specific target areas; (2) performing multiple actions seamlessly, each requiring different force application policies; and (3) motion adaptation under uncertain positional conditions. 
To address these, we propose a deep neural network (DNN)-based architecture utilizing proprioceptive and visual attention mechanisms, along with impedance control to regulate the robot’s movements.
Using the dual-arm humanoid robot Dry-AIREC, the proposed model successfully generated motions to insert the robot's hand between the bed and a mannequin's back without applying excessive force, and it supported the transition from a supine to a lifted-up position.
The project page is here: \url{https://sites.google.com/view/caregiving-robot-airec/repositioning}
\end{abstract}

\section{Introduction}
\begin{figure}[]
         \vspace*{0.2cm}
      \centering
   \includegraphics[width=6cm]{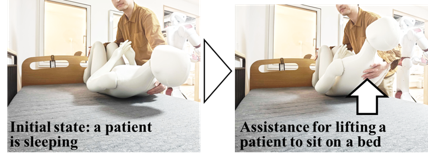} 
  \vspace*{-0.2cm}
   \caption{An example of repositioning caring tasks: a caregiver helps a patient change a posture from a sleeping state to a sitting state.}
      \label{fig:reposition}
       \vspace*{-0.6cm}
   \end{figure}

One of the key roles envisioned for humanoid robots is transforming human caregiving. The demographic shift in many countries has resulted in both a growing demand and shortage of caregiving services. Robots offer immense potential to provide physical assistance to humans through their versatility and adaptability. 
Despite this pressing demand, the practical implementation of robotics in caregiving remains limited due to ongoing technological challenges. 
Among caregiving tasks, repositioning (Fig. \ref{fig:reposition}) plays a crucial role in maintaining the health and quality of life of individuals with limited mobility \cite{gillespie2020repositioning}. 
For this reason, we focus on repositioning care as a primary application in this research.

\thispagestyle{firstpage}

In repositioning, it is common for dual-arm robots to manually mimic human motions of holding and carrying patients to achieve multiple transferring tasks \cite{cheng2024overview}. Repositioning care motion by a robot has been achieved only under human guidance \cite{brinkmann2022providing}.
Applying appropriate force to an appropriate position of a patient adapting to a situation is still challenging. 
By leveraging DNNs, we aim to avoid reliance on pre-programmed robot motions, which are inadequate for handling the complex and unpredictable conditions common in caregiving tasks.
Humanoid robots, in particular, offer significant potential for utilizing DNNs due to their high redundancy, which allows them to manage various tasks and adapt to diverse situations~\cite{suzuki2023deep}.

Previous research utilizing DNNs has demonstrated the generation of robot arm trajectories while maintaining appropriate contact forces with objects.
For instance, \cite{adachi2018imitation} achieved line drawing along a ruler, and \cite{saito2021utilization, saito2020wiping} implemented wiping motions on 3D surfaces. 
However, in caregiving tasks, simply maintaining a constant level of force is insufficient.
It is essential to: (1) determine where to apply force, avoid applying force to non-target areas, and ensure that the appropriate force is applied to the target regions; (2) smoothly transition tasks—such as inserting the robot's hand between a bed and a patient's back and then lifting them up to change the position on the bed—while recognizing and assessing the patient's state in real time; and (3) adapt to the uncertain positional conditions where occlusion sometimes occurs, commonly found in domestic caregiving settings.

To address these challenges, we construct our DNN model based on previous model, EIPL \cite{suzuki2023deep}, which integrates visual attention mechanisms, and we uniquely introduce proprioceptive attention to enhance task recognition and adaptability.
A DNN with visual attention mechanisms focuses on the key areas in images—regions rich in visually significant information.
This approach allows the robot to efficiently recognize and respond to both its target and the environment.
Additionally, we hypothesize that proprioceptive information is crucial for efficient task recognition since each action in caregiving requires distinct contact approaches, and the robot must adjust its movements based on applied force and ongoing changes. 
Proprioceptive feedback also mitigates occlusion issues, ensuring consistent perception during tasks.
We control the robot's arms using impedance control, dynamically adjusting joint softness to regulate applied force.
To best of our knowledge, this is the first autonomous supine-to-sitting repositioning motion generation by a humanoid robot. 

\section{Related Work}
This study focuses on physically assistive robots, that have the potential to reduce caregiver workload or autonomously perform caregiving tasks~\cite{maalouf2018robotics, cheng2024overview}.

\subsection{Robotic Systems for Transfer Assistance}
Robotic systems based on welfare equipment, such as a walker, provide effective assistance during patient transfers \cite{huang2021proxy, liu2022mechanics}. 
Besides, soft actuators have been developed for safer manipulation in transfer assistance, improving both patient comfort and caregiver ease \cite{choi2020development}.
Wearable robots have also been explored for caregiving support. 
Assistive suits such as HAL \cite{sankai2011hal, von2019effectiveness} reduce lower back strain for caregivers, while wearable robotic systems for patients can capture biosignals, estimate movement intentions, and assist with motion \cite{ando2010intelligent}. 
Additionally, wearable robotic limbs have demonstrated improved stability in sit-to-stand transfers \cite{sugiura2023variable}. 
However, these systems require manual setup for motions and force interactions, typically under therapist guidance, limiting their usability in real-world caregiving settings.

\subsection{Robot Arms for Caregiving Tasks using Force Interaction}
Several robotic arms have been designed to apply interaction forces to a target for caregiving tasks. 
Rule-based assistive robots have developed to work alongside caregivers for repositioning, effectively reducing caregiver workload \cite{brinkmann2022providing, kowalski2023rule}.
Other systems, such as RIBA, incorporate human-like arms designed for physically demanding tasks, enabling robots to lift and transfer patients between beds and wheelchairs \cite{mukai2010riba}. 
Similarly, RoNa demonstrated the ability to lift patients weighing up to 227 kg \cite{ding2014giving}. 
Some studies have also evaluated patient comfort by analyzing interface pressure and optimal positioning during lifting tasks \cite{ding2012comfort, guan2023analysis}.
However, these systems lack autonomous motion planning, requiring caregivers to manually adjust motion parameters such as position, direction, and speed. 
This reliance on human intervention limits their practicality in dynamic caregiving environments.

\subsection{Autonomous Manipulation for Caregiving}
Autonomous motion generation is essential for performing daily caregiving tasks. 
Skeleton recognition-based systems have been developed to detect human body-segment positions in 3D and autonomously plan reaching trajectories \cite{miyake2022skeleton}. 
However, image-based approaches for contact point identification face challenges, particularly when dealing with occluded body parts, such as the back, where robot hands may also be obscured.

In the assistive dressing scenario \cite{koganti2017bayesian, zhang2020learning, jevtic2019personalized, zhang2022learning, sun2024force} or assistive feeding scenario \cite{park2020active, gordon2023towards,nanavati2023design}, robots have addressed occlusion and variability issues using multimodal information and successfully performed autonomous arm trajectory planning around the body using Bayesian models and deep learning techniques. 
These systems can also be integrated with language models to enable adaptive behavior based on individual user requests \cite{miyake2023feasibility, jevtic2019personalized}.
However, the dressing-assistance robots primarily interact with clothing rather than making direct physical contact with the patient.
Similarly, feeding assistance robots do not involve direct physical contact with the patient \cite{park2020active,gordon2023towards,nanavati2023design}.
In contrast, repositioning tasks require direct physical interaction with the patient, demanding precise force control while avoiding non-target areas—an inherently more complex and challenging problem.


\section{Objectives and Approach}
Repositioning involves two key tasks: reaching (extending the arm to the target point) and lifting (assisting in raising the upper body). 
Each task requires distinct force application strategies.
During reaching, even a small force applied to non-target areas (e.g., the shoulder or head) can cause unintended shifts in the target’s position, and make the target uncomfortable.
In addition, the motion requires the insertion of the hand to the narrow space between the bed and the target's back, moving the hand along the surface of the bed, that needs flexible motion.
In contrast, lifting requires applying the correct force direction and magnitude to achieve proper repositioning; otherwise, insufficient or incorrect force results in no change in posture.
Additionally, reaching behind the target causes occlusions, complicating motion execution.
Our proposal model integrates visual and proprioceptive attention mechanism, which dynamically adjusts the focus area of vision, joint angles and torques, enabling automatic switching between force application and non-application policies.


   
    \begin{figure*}[]
      \centering
   \includegraphics[width=12cm]{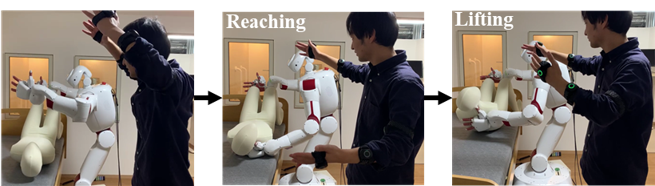}   
   \caption{Teleoperating Dry-AIREC with the motion capture system for collecting demonstration dataset}
      \label{fig:teaching}
    \vspace*{-0.1cm}
   \end{figure*}

  \begin{figure}[]
      \centering
   \includegraphics[width=8cm]{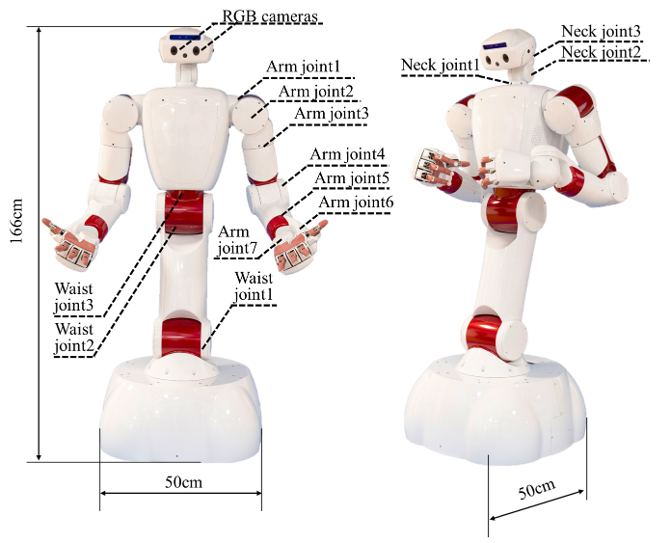}      
   \caption{Specification of the humanoid robot Dry-AIREC.}
      \label{fig:AIREC}
   \end{figure}

\section{Method}
To build and train our multimodal DNN model, we first need to collect a training dataset that includes visual data, torques, and joint angles, for this, we use a teleoperation method. 
However, precisely controlling the robot arm to reach behind the patient's head without collision, move smoothly along the bed’s surface, and apply force to the patient’s back by teleoperation is challenging. 
To facilitate this process, we use different impedance parameters for each phase—reaching and lifting—to improve manipulation accuracy. 
When it comes to motion generation, we do not need to adjust the impedance parameters since the DNN model must have both policies of reaching and lifting phases based on visuo-proprioceptive information.

\subsection{Collecting Training Data with Teleoperation}
We collect sequential vision, torque, and joint angle data of the robot by teleoperating it via a motion capture system with IMU (Perception Neuron, Noitom, Miami, FL, USA). 
The robot’s upper arms, forearms, and hands are synchronized with the operator’s movements.

As shown in Fig.~\ref{fig:teaching}, the robot first starts from an initial position with both arms raised above the caregiving target. 
Next, the robot reaches its hands to target points: the right hand stabilizes the patient's legs, while the left hand is inserted between the neck and the bed. 
Finally, the robot lifts the patient's upper body using its left hand.

To adapt to different movement phases, we adjust impedance parameters during teleoperation.
During reaching, the stiffness is set soft to limit contact force, which ease to control robot hand to move along bed surface and insert the hand between narrow space beneath the patient's neck. 
During lifting, the stiffness is set hard, allowing effective force application for strongly assisting to lift up the patient.

\subsection{Data Processing and Training}
\label{sec:data_processing}
Due to impedance control, recorded joint angles often deviate from commanded angles. 
This deviation is particularly large during lifting, where a large amount of contact introduces greater discrepancies.
To ensure consistency between reaching and lifting phases, we apply a compensation process for left-arm positions based on torque values. 
Specifically, the torque value is multiplied by 0.001 (the ratio of angle deviation caused by impedance control) for joints 1, 4, and 6 in Fig.~\ref{fig:AIREC}, which are critical joints for lifting.

After compensation, joint angles and torque values are normalized by min-max scaling. 
RGB image values were normalized by dividing by 255.
Using this processed data, we trained our DNN model to recursively predict the next sensorimotor state. 
Details of the model architecture are discussed in the next section.

\subsection{Evaluation}
To evaluate our approach, we test motion generation for both reaching and lifting while setting the lifting-phase impedance parameters for all sequential movements.
We assess the model’s ability to generalize across environments by conducting tests under six different bed heights, including three unseen conditions.
Since the wrist joint (joint 6) torque is a key indicator of contact force, we define success as follows:
\begin{itemize}
   \item Successful reaching: Less than 1 Nm torque at wrist joint, avoiding unintended contact with non-target parts such as patient's head, and contact to patient's back with its left hand and patient's knee with right hand.
   \item Successful lifting: More than 1 Nm torque at wrist joint, achieving the patient's posture change.
\end{itemize}

To assess the impact of proprioceptive attention, we compare our DNN model with proprioceptive attention against the conventional DNN model (EIPL) which does not have proprioceptive attention. 
Additionally, we analyze the latent values of visual and proprioceptive attention to better understand their contributions.

\begin{figure*}[]
   \vspace*{0.2cm}
      \centering
      \includegraphics[width=17.5cm]{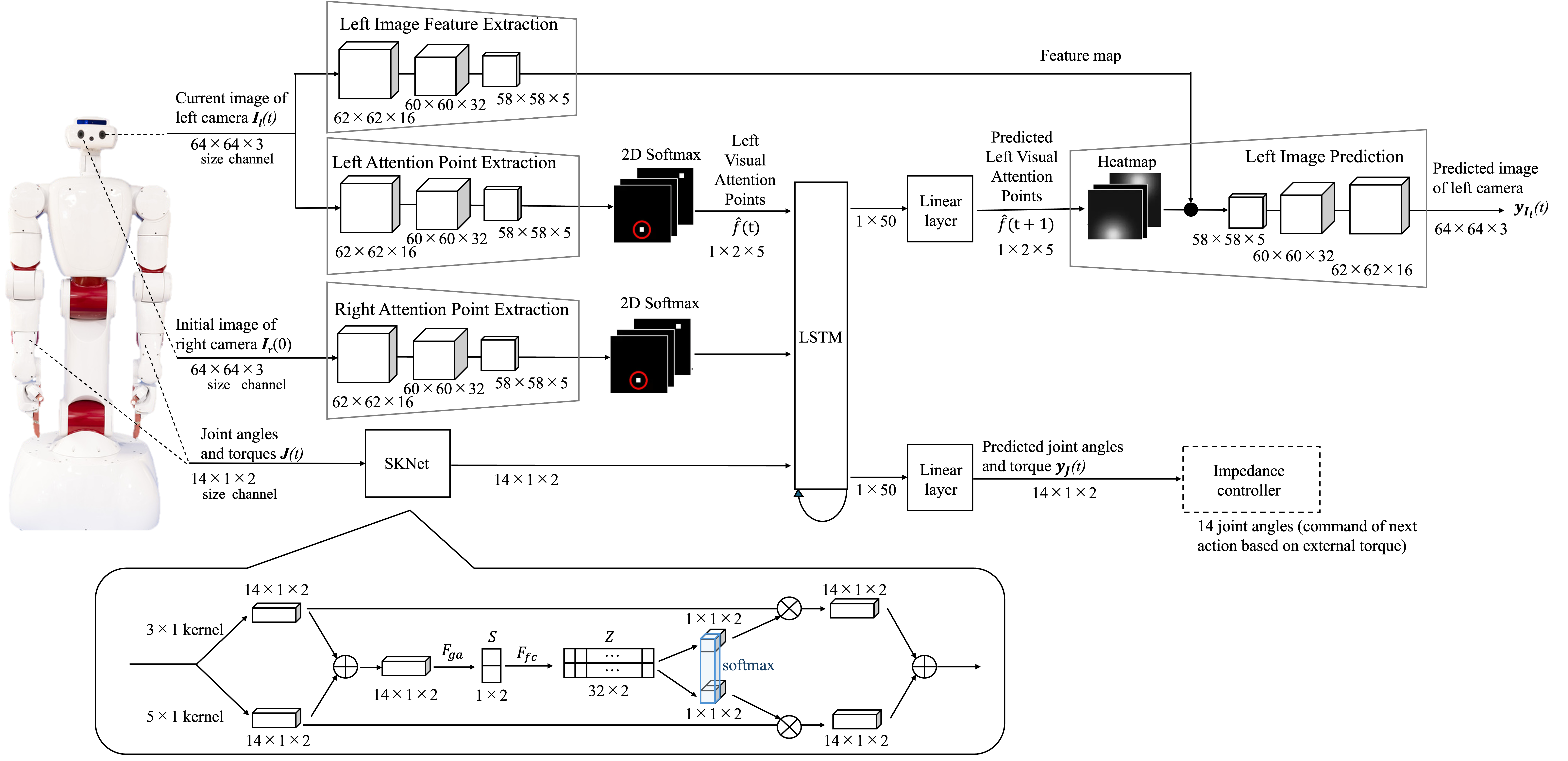}
      \caption{Overview of the proposed DNN architecture, based on EIPL with visual attention, which we add proprioceptive attention mechanism.}
      \label{fig::algorithm}
    \vspace*{-0.3cm}      
\end{figure*}
\section{DNN model}
Fig. \ref{fig::algorithm} shows the overall architecture and the parameters of the proposed model, which builds upon the deep predictive learning model EIPL \cite{suzuki2023deep}. 
The left eye camera image $I_l(t)$, the initial step of the right eye camera image $I_r(0)$, and torque sensor and joint data $J(t)$ are input to the model, where $t$ denotes the time step.
The model, based on long short-term memory (LSTM), learns to predict the next-step left image $I_l(t+1)$ and joint torque/angle $J(t+1)$, minimizing the prediction error $E$: 
\begin{equation}
E =\sum_{t}{(y_{I_l}(t)-I_l(t+1))^2+(y_{J}(t)-J(t+1))^2}+E_a
\label{y}
\end{equation}
where $y$ presents the model output and $E_a$ is the visual attention error, described later.
By predicting both next-step image and torque data alongside joint angles, the model prevents excessive reliance on joint states alone.
When the robot executes a motion for the test, each sensorimotor data $(I_l(t), I_r(0), J(t))$ is input to the model, and the predicted joint angle data $y_{\rm {J}}(t)$ is used for the robot control ($\therefore J(t+1) = y_{\rm {J}}(t) $).

\subsection{Proprioceptive attention}
To integrate proprioceptive attention into EIPL, we use a Selective Kernel Network (SKNet) \cite{li2019selective}, allowing dynamic adaptation of joint and torque feature selection.
Human proprioceptive attention mechanisms dynamically adjust neural populations encoding relevant task features \cite{gomez2016neural}.
Inspired by the human mechanism, we propose proprioceptive attention to distinguish proprioceptive states (i.e. joint angles and torques) for the smooth transition between different force interaction policies.
SKNet is a CNN architecture designed to enhance feature selection and representation learning, which is inspired by the adaptive receptive field size of neurons in the human brain.
It employs multiple parallel branches with different kernel sizes, allowing the network to capture diverse feature representations. 
By dynamically selecting the most relevant kernel for each input, SKNet adapts to task-specific variations, improving the effectiveness of learned representations.

In our model, $14\times1\times2$ (14 joints for angle/torque channels) data is passed through parallel convolution layers with kernel sizes $3\times1$ and $5\times1$ to extract features.
The tensors \textbf{\textit{a}} and \textbf{\textit{b}} described in Fig. \ref{fig::algorithm}, which are latent values, represent the attention weights corresponding to the first branch (corresponding to the $3\times1$-kernel convolution) and the second branch (corresponding to the $5\times1$-kernel convolution), respectively. 
These attention weights ($a, b$) are calculated using softmax function based on the the fully connected layers outputs ($F_{fc}$) applied to the global average pooled feature representation ($F_{ga}$). 
The sum of values in each channel of \textbf{\textit{a}} and \textbf{\textit{b}} is 1.0, ensuring adaptive feature selection.
Through LSTM and the linear layer, the next-step joint torque and angles are predicted and sent to the robot's impedance controller as a command of the next motion $J(t+1)$.

\subsection{Visual attention}
We implement visual attention based on Yasutomi et al. \cite{yasutomi2023visual} and EIPL \cite{suzuki2023deep}.
The model extracts spatially important features from the left-eye camera image ($64\times64\times3$ ($W\times H\times C$)) using convolutional neural network (CNN) modules.
All of them use the $3\times3$ kernels, 1 stride and 0 padding, and each module has its own set of weights.
At first, the Left Feature Extraction module extracts low-dimensional image feature maps of the input image.
Simultaneously, the Left Attention Point Extraction module applies spatial softmax to identify the key locations and outputs the positional coordinates in the image $\hat{f}(t)$ (called the visual attention points in this paper) from the input image.
Next, LSTM and linear layer predict the next time attention points $\hat{f}(t+1)$, and then it can generate a heatmap with inverse spatial softmax.
Then the Left Image Prediction module predicts the original-dimensional image at the next time t+1 on the basis of the predicted heatmap and the low-dimensional feature map obtained from the Left Image Feature Extraction module.
We use both image feature and heatmap because the heatmap alone is not sufficient to predict the image, but the image can be easily predicted by using the information near the attention points among the image features obtained from the Left Image Feature Extraction module.

We calculate the visual attention error $E_a$ as,
\begin{equation}
E_a =\frac{1}{K}||\hat{f}(t)-\hat{f}(t+1)||_{2}^2,
\label{E_a}
\end{equation}
where $K=1\times2\times5$, as we track five attention points in a 2D image space.
It denotes the error in the Euclidean distance between the attention points of the encoder output $\hat{f}(t)$ and the decoder input $\hat{f}(t+1)$. 
By adding $E_a$ to the error function $E$, the predicted visual attention points can be efficiently searched near the current visual attention points. 
Since visual attention points track the robot’s hand and target objects, their movement is not significant per time step, making Euclidean distance an effective measure of prediction accuracy.

Additionally, the initial image of the right camera $I_r(0)$ is used in this study.
The architecture of the Right Attention Point Extraction module is the same as the left one.
The initial step of the right image is used to support the model for predicting the next sensorimotor data better by consistently recalling the environmental information, especially depth information by adding a different camera view.

\subsection{Training setup}
The activation function we use is leaky ReLU.
The DNN model was trained, and the 9000th epoch was selected for testing as it minimized total validation error.

Since all modules are end-to-end connected and the proprioceptive attention weights and visual attention points are self-organized, we do not need explicit training labels. 
Instead, the DNN model learns the attention mechanisms in a way that optimally reduces prediction error, ensuring that the learned proprioceptive attention weights and visual attention points align with task-relevant features.

\section{Experimental Setup}

\subsection{Robot constitutions}
We utilized the Dry-AIREC humanoid robot (Fig.~\ref{fig:AIREC}), developed by Tokyo Robotics Inc. based on Torobo \cite{TokyoRobot}. 
This robot was chosen for its impedance control capabilities, allowing customization of mass, damping, and spring parameters. 
The compliance control mode enables dynamic joint stiffness adjustment, improving adaptability and ensuring safer interactions with humans and environment.
To enhance grip and tactile feedback, the robot's fingertips and palms are covered with a rubber material. 
We controlled its dual 7-DOF arms and utilized both eye cameras and torque sensors on all arm joints for motion execution and feedback, wheres we fixed head and torso joint angles.

\subsection{Environment and Dataset}
In this study, we used a mannequin SM0531 (Shirai Ltd., Japan) as the caregiving target. 
The mannequin measured 1800 mm in height, 470 mm in shoulder width, 885 mm in leg length, and 750 mm in arm length, and was made of urethane. 
During the experiment, it lay on its back with bent legs. 
The mannequin required approximately 72 N of force to adjust its posture, though its light weight (8 kg) made it highly susceptible to unintentional shifts even with small external forces (around 1 N). 
This necessitated careful control of the robot’s applied force and direction during repositioning tasks.

The bed used in this study measured 205 cm in width and 95 cm in depth, with an adjustable height. 
To train the DNN model, data was collected at bed heights of 59 cm, 65 cm, and 71 cm.
For evaluation, additional heights of 56 cm, 62 cm, and 68 cm were used to test the model’s adaptability in unknown conditions.
These variations required the robot to dynamically generate adaptive arm trajectories based on spatial recognition of the patient and environment.

We collected teleoperated data three times each under three conditions of bed heights. 
In total, nine datasets of images and robot states during dual-arm movements were collected as training data.
The sampling frequency of the algorithm was 20 Hz. The number of time step for one dataset was 260.

\subsection{Impedance Control Parameters} 
The impedance control parameters of each joint for both hard and soft modes were heuristically determined. 
In the soft mode, the robot hands moved 1 cm when 3 N external force was applied, wheres in the hard mode, the same displacement required 8 N force. 
During training data collection via teleoperation, soft mode was applied for the reaching phase to minimize unintended force, while hard mode was used for the lifting phase to ensure effective assistance.
The mode switching was manually controlled via keyboard input. 
On the other hand, we used hard modes for all the motions during evaluation.
The compensation process addressing the discrepancy in reaching motion is detailed in Section~\ref{sec:data_processing}.
%

\begin{table}[]
 \vspace*{0.1cm}
\caption{Success rates of motion generation without and with SKNet.}
\label{table_example}
\begin{center}
\begin{tabular}{|c|c|c|c|c|}
\hline
\multirow{2}{*}{Condition} &  \multicolumn{2}{|c|}{ Without SKNet (baseline)} & \multicolumn{2}{|c|}{ With SKNet (ours)}\\
\cline{2-2} \cline{3-5}
& Reaching & Lifting & Reaching & Lifting\\

\hline
Known & 0/30 & N & 30/30 & 30/30 \\
\hline
Unknown & 0/30 & N & 29/30 & 27/30 \\
\hline
\end{tabular}
\end{center}
      \label{tab:success}
\end{table}

  \begin{figure}[]
      \centering
   \includegraphics[width=8.5cm]{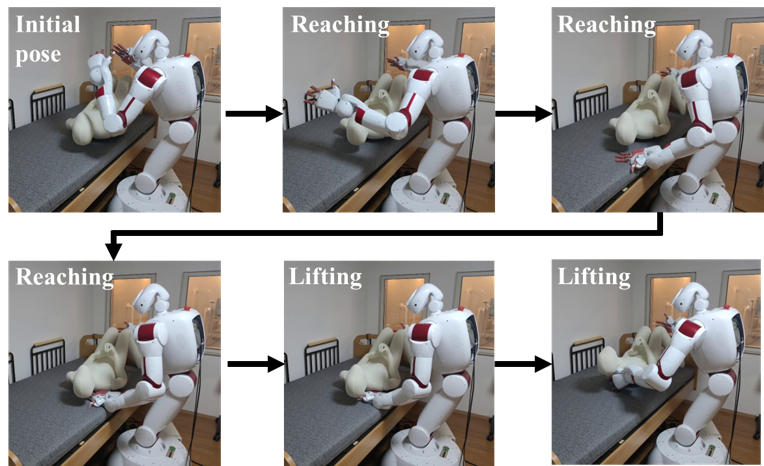}        
   \caption{Scenes of the test of the motion generation with the proposed architecture in a condition of untrained bed height (67cm).}
      \label{fig:testmotion}
   \end{figure}

  \begin{figure}[]
      \centering
   \includegraphics[width=5cm]{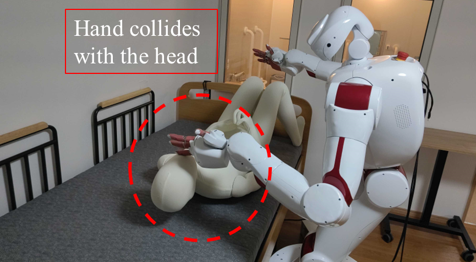} 
   \caption{An example of a failed scene of the reaching motion with baseline model. The robot hit the patient's head and could not insert its hand beneath the patient's neck.}
      \label{fig:failed}
   \end{figure}

\section{Results and Discussion}
\subsection{Motion Generation}
Table \ref{tab:success} shows the success rates of the motion generation by the systems without and with SKNet.
The model with SKNet (ours) resulted in a high success rate of completing the motion even in unknown conditions.
Fig. \ref{fig:testmotion} shows Dry-AIREC successfully executed repositioning motions using the proposed architecture, even at an untrained bed height (67cm). 
The model achieved 100\% success rate (30/30 trials) at trained bed heights and 90\% success rate (27/30 trials) at untrained heights. 
In these trials, the robot successfully extended its hand to the back of the mannequin without excessive contact while providing sufficient support for postural adjustment.
In contrast, the baseline model without SKNet (EIPL) consistently failed to complete the motion, as it frequently made unintended contact with the mannequin's head (Fig. \ref{fig:failed}).
Without SKNet, the system was unable to properly distinguish between the target regions, leading to confusion between collision avoidance and the application of supportive force. 
These results confirm that proprioceptive attention using SKNet is essential for generating effective repositioning motions.

  \begin{figure}[]
      \centering
   \includegraphics[width=6.5cm]{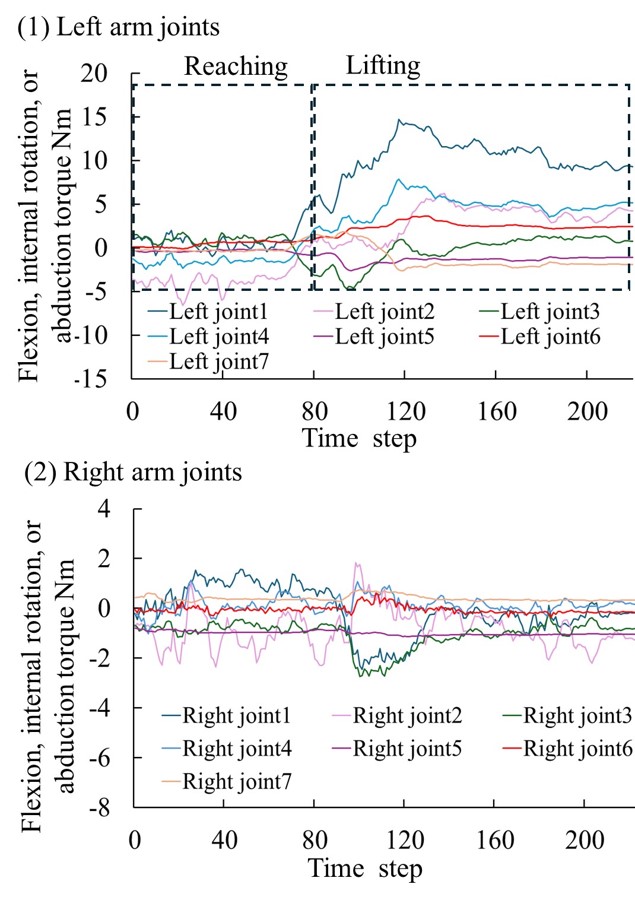} 
   \caption{Examples of time-series external torque data applied to the mannequin.
   }
      \label{fig:torquetime}
   \end{figure}

  \begin{figure}[]
      \centering
   \includegraphics[width=8cm]{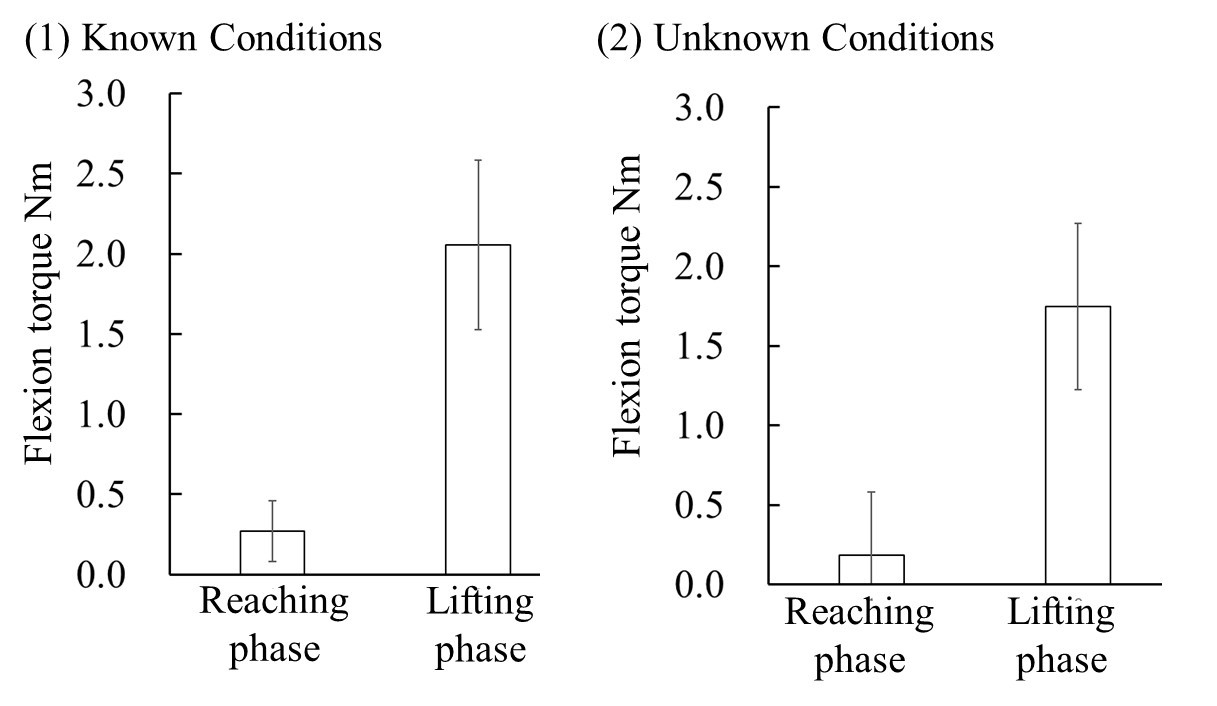} 
   \caption{Mean and standard deviation of external torque data of the left wrist joint for the known and unknown conditions.}
      \label{fig:torquescene}
   \end{figure}

To further evaluate the system’s performance, we analyzed the torque measurements during repositioning with the proposed DNN model.
Fig. \ref{fig:torquetime} shows time-series external torque data, where the positive direction of torque corresponds to the force applied to the mannequin during the lifting phase. 
The right joint 6 exhibited increased torque at the start of the lifting phase, indicating that the right arm firmly stabilized the mannequin’s legs in coordination with the left arm’s movement.
Fig. \ref{fig:torquescene} displays the mean and standard deviation of torque values of left wrist joint 6 during the reaching and lifting phases.
The data indicate a clear phase-dependent torque application, with minimal left-arm torque less than 1 Nm during reaching and appropriately increased torque more than 1 Nm during lifting. 
This confirms that the model effectively distinguishes between motion phases and regulates force application accordingly.

\begin{figure}[]
      \centering
   \includegraphics[width=6.6cm]{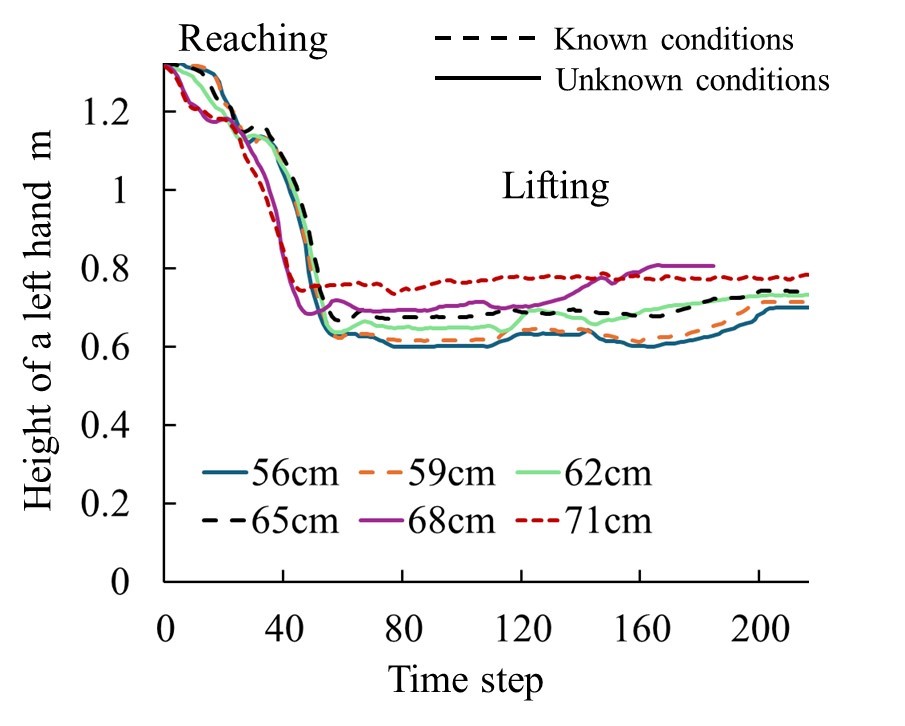} 
      \vspace*{-0.3cm}
   \caption{Trajectory of height of the left hand.}
      \label{fig:tftrajectory}
      \vspace*{-0.1cm}
   \end{figure}

Fig. \ref{fig:tftrajectory} presents time-series trajectories of Dry-AIREC’s left-hand (palm) height across different bed height conditions, with the proposed DNN model.
The results show that the robot’s hand successfully adjusted its target height based on bed height variations, demonstrating adaptability in both known and unknown conditions.
The velocity of the motion and the timing of the starting lifting motion varied among trials and conditions. We believe that the variability appeared as a result of the model adjusting the generation of the next-step motion corresponding to the situation/scene dynamically.

\subsection{Proprioceptive attention}
\begin{figure}[]
      \centering
   \includegraphics[width=8cm]{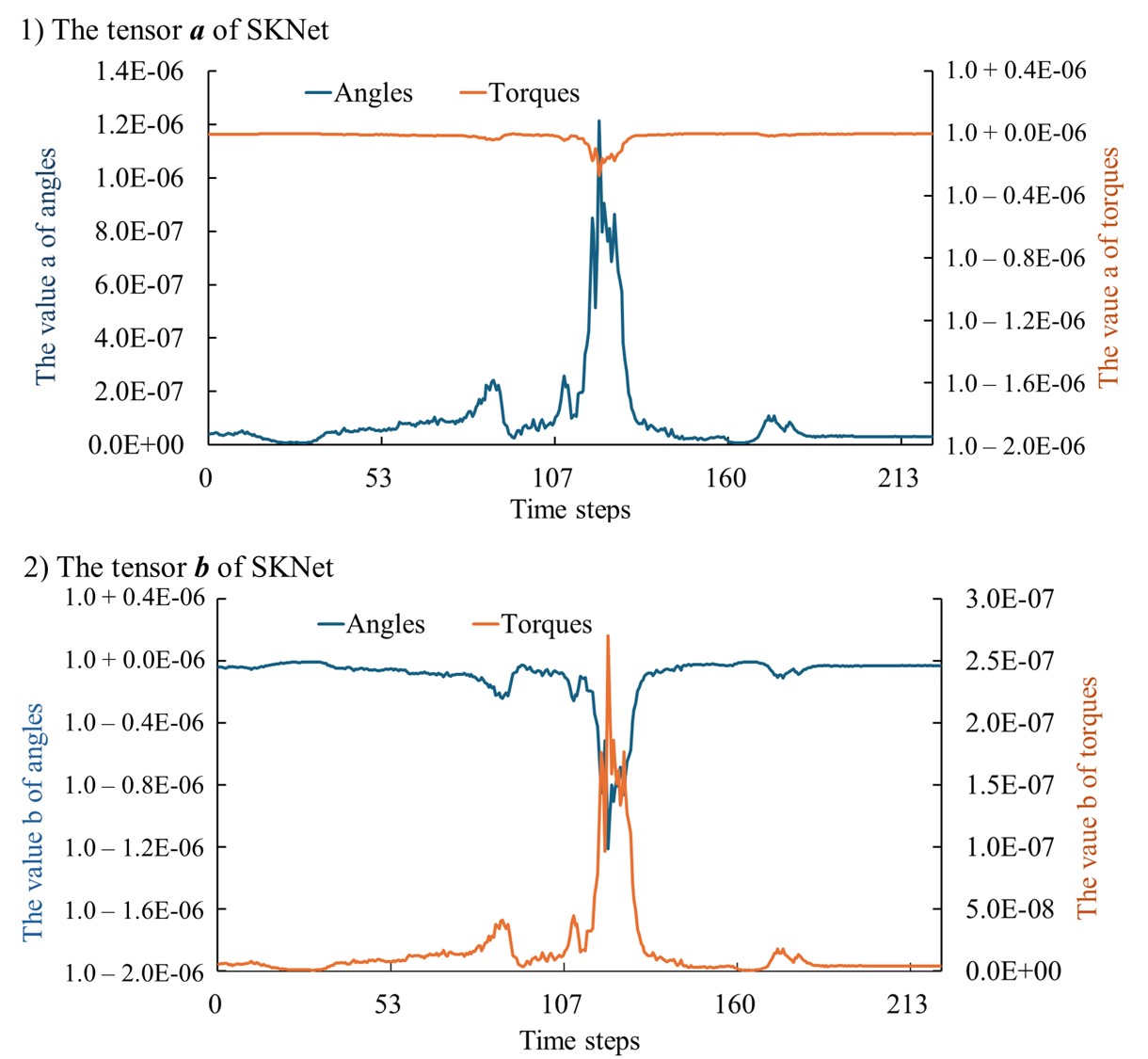} 
   \caption{Latent values of the SKNet. The latent value \textbf{\textit{a}} is correspond to the 3x1-kernel CNN, and  \textbf{\textit{b}} is correspond to the 5x1-kernel CNN. }
      \label{fig:latent_skn}
   \end{figure}

We analyze the proposed model's attention mechanism to verify its effectiveness in motion generation.
Fig. \ref{fig:latent_skn} shows the sequential flow of the latent value, proprioceptive attention weight (\textbf{\textit{a}} and \textbf{\textit{b}}) of the SKNet.
The tensor \textbf{\textit{a}} corresponds to the latent value associated with the $3\times1$-kernel convolution, which captures features from a smaller region, and \textbf{\textit{b}} corresponds to the $5\times1$-kernel convolution, which extracts features from a larger region. 
These values indicate the appropriate kernel size for predicting next-step sensorimotor data depending on the situation.
In other words, they reveal whether the model prioritizes fine-grained local features or broader contextual information in different modalities.

In most cases, the value of \textbf{\textit{a}} was close to 0.0 while \textbf{\textit{b}} was near 1.0 for joint angles, whereas the opposite pattern was observed for torques, with \textbf{\textit{a}} around 1.0 and \textbf{\textit{b}} close to 0.0. 
This suggests that SKNet learned to focus on large-region features for joint angles and small-region features for torques.
We infer that large-region features were necessary for angles because the larger number of joints were coordinated to perform the repositioning task.
Conversely, small-region features were preferred for torque data since only a few joints were mainly involved in supporting the mannequin’s weight, as shown in Fig. \ref{fig:torquetime}.

However, we observed fluctuations in \textbf{\textit{a}} and \textbf{\textit{b}} after a stronger torque was applied — when the lifting motion began. 
This suggests that proprioceptive attention contributes to distinguish the timing to start lifting, and realize a seamless transition between task phases and force application policies.
Therefore, dynamically adjusting the kernel size in the attention mechanism based on the proprioceptive state was essential for effective motion generation.

Additionally, we experimented with training the model using data from only a single bed height.
In this case, the baseline model (without SKNet) was still able to reach the mannequin while avoiding contact with its head.
However, when tested on different bed heights, the baseline model consistently failed.
This confirms that proprioceptive information must be transformed into more interpretable features to ensure robust and adaptable motion generation across different environments.

\subsection{Visual attention}
\begin{figure}[]
      \centering
   \includegraphics[width=8.5cm]{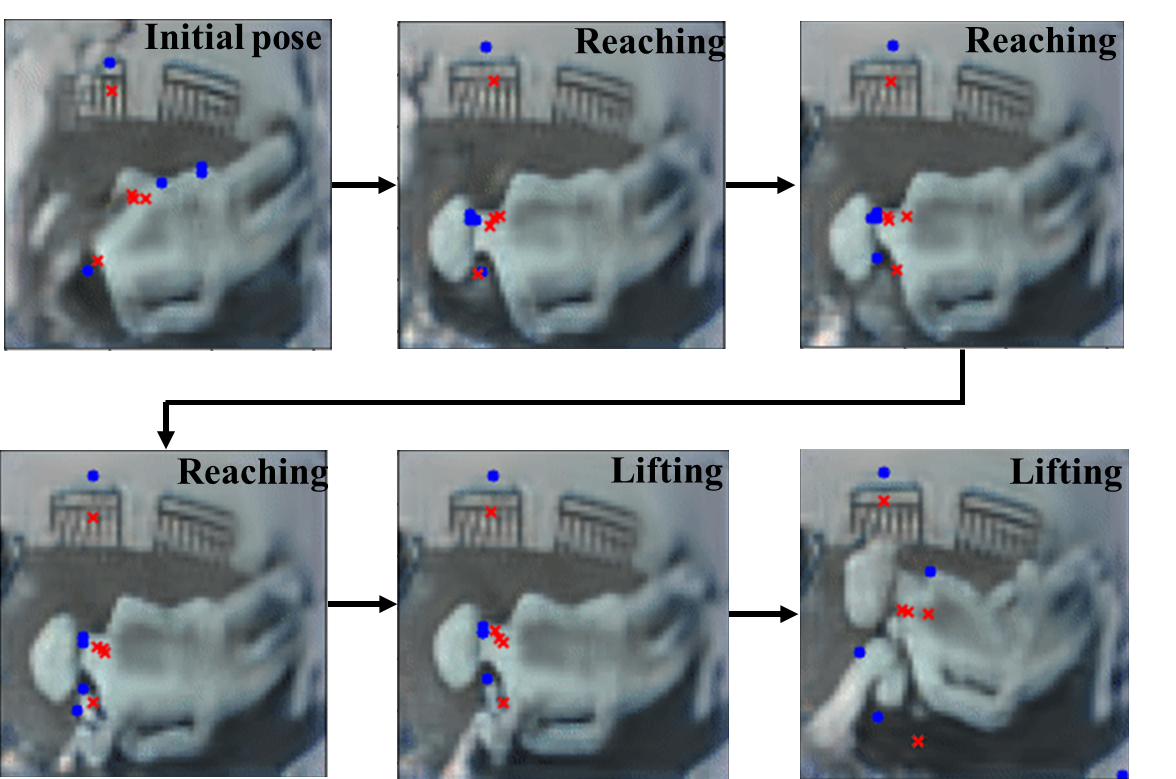} 
   \caption{Scenes of the predicted images with attention points.
   Blue circle points and red x-shape points indicate current and future image attention, respectively.}
      \label{fig:image_attn}
   \end{figure}

Fig. \ref{fig:image_attn} shows an example of predicted images with attention point visualizations.
The blue circle points and the red x-shape points indicate current and predicted attention points ($\hat{f}(t)$ and $\hat{f}(t+1)$), respectively.
The attention mechanism focused not only on the mannequin's neck region but also on the left handrail of the bed. 
Since the robot's dual-arm control relied on spatial information from both the mannequin and the bed, the arm-joint trajectories were able to adapt accordingly to different bed heights.

Notably, attention was concentrated around Dry-AIREC’s hand, particularly at the lower left and center of Fig. \ref{fig:image_attn}.
The lower-center image captures the critical moment when the motion phase transitions from reaching to lifting.  
This suggests that the visual attention mechanism plays a crucial role in detecting phase transitions and coordinating vision-proprioceptive integration for smooth and adaptive motion execution.

\section{Conclusion}
We proposed a deep predictive learning architecture integrating proprioceptive and visual attention to enable a humanoid robot to assist in repositioning care, which require the robot to (1) apply appropriate force to specific target areas; (2) perform multiple actions seamlessly, each requiring different force application policies; and (3) adapt to changes in the positional conditions.  
Using the Dry-AIREC humanoid robot, which allows joint stiffness adjustment for impedance control, we conducted experiments.
The proposed DNN model successfully guided the robot’s hand to reach the back of the mannequin while minimizing excessive contact forces, and then provide appropriate support for postural adjustment during lifting.
Our results demonstrated that integrating visual attention with EIPL and proprioceptive attention through SKNet significantly improved motion adaptability and success rates, even in untrained conditions.
Analysis of SKNet attention weights confirmed that the model effectively distinguished between large- and small-region feature extraction depending on the sensory modality (joint torque and angle) and task phase.
Additionally, the visual attention mechanism focused on key areas, such as the mannequin’s neck, the robot’s hand, and the bed handrail, enabling robust spatial representation and environmental adaptation.

For future work, several extensions will be considered.
First, incorporating torso motion alongside dual-arm control could enhance the robot’s range of motion and dexterity.
Second, expanding the training dataset to include variations in target shape, size, texture, and color would improve generalization to diverse real-world conditions.
Third, given the limited field of view of Dry-AIREC’s cameras, active perception through neck movement could be explored to optimize visual input dynamically.
By further developing these aspects, we aim to enhance the adaptability and effectiveness of humanoid robots in real-world caregiving scenarios.

\bibliographystyle{ieeetr}
\bibliography{root}

\end{document}